\documentclass{article} 
\usepackage{iclr2017_conference,times}
\usepackage{hyperref}
\usepackage{url}
\usepackage{epsfig}
\usepackage{graphicx}
\usepackage{amsmath}
\usepackage{amssymb}

\usepackage{algorithm}
\usepackage{algorithmic}
\usepackage{stfloats}
\usepackage{amsfonts}
\usepackage{slashbox}
\usepackage{pdfpages}

\usepackage{multirow}

\usepackage{subfigure}

\title{Boosting Image Captioning with Attributes}

\author{Ting Yao, Yingwei Pan, Yehao Li, Zhaofan Qiu, Tao Mei\\
Microsoft Research Asia\\
\texttt{\{tiyao, v-yipan, v-yehl, v-zhqiu, tmei\}@microsoft.com} \\
}

\begin{document}

\maketitle

\begin{abstract}
Automatically describing an image with a natural language has been an emerging challenge in both fields of computer vision and natural language processing. In this paper, we present Long Short-Term Memory with Attributes (LSTM-A) - a novel architecture that integrates attributes into the successful Convolutional Neural Networks (CNNs) plus Recurrent Neural Networks (RNNs) image captioning framework, by training them in an end-to-end manner. To incorporate attributes, we construct variants of architectures by feeding image representations and attributes into RNNs in different ways to explore the mutual but also fuzzy relationship between them. Extensive experiments are conducted on COCO image captioning dataset and our framework achieves superior results when compared to state-of-the-art deep models. Most remarkably, we obtain METEOR/CIDEr-D of 25.2\%/98.6\% on testing data of widely used and publicly available splits in \citep{Karpathy:CVPR15} when extracting image representations by GoogleNet and achieve to date top-1 performance on COCO captioning Leaderboard.
\end{abstract}

\section{Introduction}
Accelerated by tremendous increase in Internet bandwidth and proliferation of sensor-rich mobile devices, image data has been generated, published and spread explosively, becoming an indispensable part of today's big data. This has encouraged the development of advanced techniques for a broad range of image understanding applications. A fundamental issue that underlies the success of these technological advances is the recognition \citep{Szegedy14,Simonyan14,He:CVPR16}. Recently, researchers have strived to automatically describe the content of an image with a complete and natural sentence, which has a great potential impact for instance on robotic vision or helping visually impaired people. Nevertheless, this problem is very challenging, as description generation model should capture not only the objects or scenes presented in the image, but also be capable of expressing how the objects/scenes relate to each other in a nature sentence.

The main inspiration of recent attempts on this problem \citep{Donahue14,Vinyals14,Xu:ICML15,You:CVPR16} are from the advances by using RNNs in machine translation \citep{Sutskever:NIPS14}, which is to translate a text from one language (e.g., English) to another (e.g., Chinese). The basic idea is to perform a sequence to sequence learning for translation, where an encoder RNN reads the input sequential sentence, one word at a time till the end of the sentence and then a decoder RNN is exploited to generate the sentence in target language, one word at each time step. Following this philosophy, it is natural to employ a CNN instead of the encoder RNN for image captioning, which is regarded as an image encoder to produce image representations.

While encouraging performances are reported, these CNN plus RNN image captioning methods translate directly from image representations to language, without explicitly taking more high-level semantic information from images into account. Furthermore, attributes are properties observed in images with rich semantic cues and have been proved to be effective in visual recognition \citep{Devi:ICCV11}. A valid question is how to incorporate high-level image attributes into CNN plus RNN image captioning architecture as complementary knowledge in addition to image representations. We investigate particularly in this paper the architectures by exploiting the mutual relationship between image representations and attributes for enhancing image description generation. More importantly, to better demonstrate the impact of simultaneously utilizing the two kinds of representations, we devise variants of architectures by feeding them into RNN in different placements and moments, e.g., leveraging only attributes, inserting image representations first and then attributes or vice versa, and inputting image representations/attributes once or at each time step.

\section{Related Work}\label{sec:RW}
The research on image captioning has proceeded along three different dimensions: template-based methods \citep{Kulkarni:PAMI13,Yang:EMNLP11,Mitchell:EACL12}, search-based approaches \citep{Farhadi:ECCV10,Ordonez:NIPS11,devlin2015language}, and language-based models \citep{Donahue14,Kiros:ICML14, Mao:NIPS14,Vinyals14,Xu:ICML15,Wu:CVPR16,You:CVPR16}.

The first direction, template-based methods, predefine the template for sentence generation which follows some specific rules of language grammar and split sentence into several parts (e.g., subject, verb, and object). With such sentence fragments, many works align each part with image content and then generate the sentence for the image. Obviously, most of them highly depend on the templates of sentence and always generate sentence with syntactical structure. For example, Kulkarni \emph{et al.} employ Conditional Random Field (CRF) model to predict labeling based on the detected objects, attributes, and prepositions, and then generate sentence with a template by filling in slots with the most likely labeling \citep{Kulkarni:PAMI13}. Similar in spirit, Yang \emph{et al.} utilize Hidden Markov Model (HMM) to select the best objects, scenes, verbs, and prepositions with the highest log-likelihood ratio for template-based sentence generation in \citep{Yang:EMNLP11}. Furthermore, the traditional simple template is extended to syntactic trees in \citep{Mitchell:EACL12} which also starts from detecting attributes from image as description anchors and then connecting ordered objects with a syntactically well-formed tree, followed by adding necessary descriptive information.

Search-based approaches ``generate" sentence for an image by selecting the most semantically similar sentences from sentence pool or directly copying sentences from other visually similar images. This direction indeed can achieve human-level descriptions as all sentences are from existing human-generated sentences. The need to collect human-generated sentences, however, makes the sentence pool hard to be scaled up. Moreover, the approaches in this dimension cannot generate novel descriptions. For instance, in \citep{Farhadi:ECCV10}, an intermediate meaning space based on the triplet of object, action, and scene is proposed to measure the similarity between image and sentence, where the top sentences are regarded as the generated sentences for the target image. Ordonez \emph{et al.} \citep{Ordonez:NIPS11} search images in a large captioned photo collection by using the combination of object, stuff, people, and scene information and transfer the associated sentences to the query image. Recently, a simple $k$-nearest neighbor retrieval model is utilized in \citep{devlin2015language} and the best or consensus caption is selected from the returned candidate captions, which even performs as well as several state-of-the-art language-based models.

Different from template-based and search-based models, language-based models aim to learn the probability distribution in the common space of visual content and textual sentence to generate novel sentences with more flexible syntactical structures. In this direction, recent works explore such probability distribution mainly using neural networks for image captioning. Kiros \emph{et al.} \citep{Kiros:ICML14} take the neural networks to generate sentence for an image by proposing a multimodal log-bilinear neural language model. In \citep{Vinyals14}, Vinyals \emph{et al.} propose an end-to-end neural networks architecture by utilizing LSTM to generate sentence for an image, which is further incorporated with attention mechanism in \citep{Xu:ICML15} to automatically focus on salient objects when generating corresponding words. More recently, in \citep{Wu:CVPR16}, high-level concepts/attributes are shown to obtain clear improvements on image captioning task when injected into existing state-of-the-art RNN-based model and such visual attributes are further utilized as semantic attention in \citep{You:CVPR16} to enhance image captioning.

In short, our work in this paper belongs to the language-based models. Different from most of the aforementioned language-based models which mainly focus on sentence generation by solely depending on image representations \citep{Donahue14,Kiros:ICML14,Mao:NIPS14,Vinyals14,Xu:ICML15} or high-level attributes \citep{Wu:CVPR16}, our work contributes by studying not only jointly exploiting image representations and attributes for image captioning, but also how the architecture can be better devised by exploring mutual relationship in between. It is also worth noting that \citep{You:CVPR16} also additionally involve attributes for image captioning. Ours is fundamentally different in the way that \citep{You:CVPR16} is as a result of utilizing attributes to model semantic attention to the locally previous words, as opposed to holistically employing attributes as a kind of complementary representations in this work.

\section{Boosting Image Captioning with Attributes}\label{sec:ICA}
In this paper, we devise our CNN plus RNN architectures to generate descriptions for images under the umbrella of additionally incorporating the detected high-level attributes. Specifically, we begin this section by presenting the problem formulation and followed by five variants of our image captioning frameworks with attributes.

\subsection{Problem Formulation}
Suppose we have an image ${I}$ to be described by a textual sentence $\mathcal {S}$, where $\mathcal{S} = \{w_1, w_2, ..., w_{N_s}\}$ consisting of $N_s$ words. Let ${\bf{I}}\in {\mathbb{R}}^{D_v}$ and ${\bf{w}}_t\in {{\mathbb{R}}^{D_s}}$ denote the $D_v$-dimensional image representations of the image ${I}$ and the $D_s$-dimensional textual features of the $t$-th word in sentence $\mathcal{S}$, respectively. Furthermore, we have feature vector ${\bf{A}}\in {\mathbb{R}}^{D_a}$ to represent the probability distribution over the high-level attributes for image ${I}$. Specifically, we train the attribute detectors by using the weakly-supervised approach of Multiple Instance Learning (MIL) in \citep{Fang:CVPR15} and treat the final image-level response probabilities of all the attributes as ${\bf{A}}$.

Inspired by the recent successes of probabilistic sequence models leveraged in statistical machine translation \citep{Bahdanau14, Sutskever:NIPS14}, we aim to formulate our image captioning models in an end-to-end fashion based on RNNs which encode the given image and/or its detected attributes into a fixed dimensional vector and then decode it to the target output sentence. Hence, the sentence generation problem we explore here can be formulated by minimizing the following energy loss function as
\begin{equation}\label{Eq:Eq5}\small
E({\bf{I}}, {\bf{A}}, {\mathcal {S}}) = -\log {\Pr{({\mathcal {S}}|{\bf{I}}, {\bf{A}})}},
\end{equation}
which is the negative log probability of the correct textual sentence given the image representations and detected attributes.

Since the model produces one word in the sentence at each time step, it is natural to apply chain rule to model the joint probability over the sequential words. Thus, the $\log$ probability of the sentence is given by the sum of the $\log$ probabilities over the word and can be expressed as
\begin{equation}\label{Eq:Eq6}\small
\log {\Pr{({\mathcal {S}}|{\bf{I}}, {\bf{A}})}} =  \sum\limits_{t = 1}^{{N_s}} {\log \Pr\left( {\left. {{{\bf{w}}_t}} \right|{\bf{I}}, {\bf{A}},{{\bf{w}}_0}, \ldots ,{{\bf{w}}_{t - 1}}} \right)}.
\end{equation}
By minimizing this loss, the contextual relationship among the words in the sentence can be guaranteed given the image and its detected attributes.

We formulate this task as a variable-length sequence to sequence problem and model the parametric distribution $\Pr\left( {\left. {{{\bf{w}}_t}} \right|{\bf{I}}, {\bf{A}},{{\bf{w}}_0}, \ldots ,{{\bf{w}}_{t - 1}}} \right)$ in Eq.(\ref{Eq:Eq6}) with Long Short-Term Memory (LSTM), which is a widely used type of RNN. The vector formulas for a LSTM layer forward pass are summarized as below. For time step $t$, ${\bf{x}}^t$ and ${\bf{h}}^t$ are the input and output vector respectively, $\bf{T}$ are input weights matrices, $\bf{R}$ are recurrent weight matrices and $\bf{b}$ are bias vectors. Sigmoid $\sigma$ and hyperbolic tangent $\phi$ are element-wise non-linear activation functions. The dot product of two vectors is denoted with $\odot$. Given inputs ${\bf{x}}^t$, ${\bf{h}}^{t-1}$ and ${\bf{c}}^{t-1}$, the LSTM unit updates for time step $t$ are:
\begin{eqnarray*}\small
~~~~{{\bf{g}}^t} = \phi ({{\bf{T}}_g}{{\bf{x}}^t} + {{\bf{R}}_g}{{\bf{h}}^{t - 1}} + {{\bf{b}}_g}),~~~~~~{{\bf{i}}^t} = \sigma ({{\bf{T}}_i}{{\bf{x}}^t} + {{\bf{R}}_i}{{\bf{h}}^{t - 1}} + {{\bf{b}}_i}),\\
~~~~~~{{\bf{f}}^t} = \sigma ({{\bf{T}}_f}{{\bf{x}}^t} + {{\bf{R}}_f}{{\bf{h}}^{t - 1}} + {{\bf{b}}_f}),~~~~~{{\bf{c}}^t} = {\bf{g}}^t \odot {\bf{i}}^t + {\bf{c}}^{t-1} \odot {\bf{f}}^t,~~~~~~~\\
~~~~~{{\bf{o}}^t} = \sigma ({{\bf{T}}_o}{{\bf{x}}^t} + {{\bf{R}}_o}{{\bf{h}}^{t - 1}} + {{\bf{b}}_o}),~~~~~~~~~~{{\bf{h}}^t} = \phi ({\bf{c}}^t) \odot {\bf{o}}^t,~~~~~~~~~~~~~~~~
\end{eqnarray*}
where ${{\bf{g}}^t}$, ${{\bf{i}}^t}$, ${{\bf{f}}^t}$, ${{\bf{c}}^t}$, ${{\bf{o}}^t}$, and ${{\bf{h}}^t}$ are cell input, input gate, forget gate, cell state, output gate, and cell output of the LSTM, respectively.

\begin{figure}[!tb]
\centering {\includegraphics[width=0.95\textwidth]{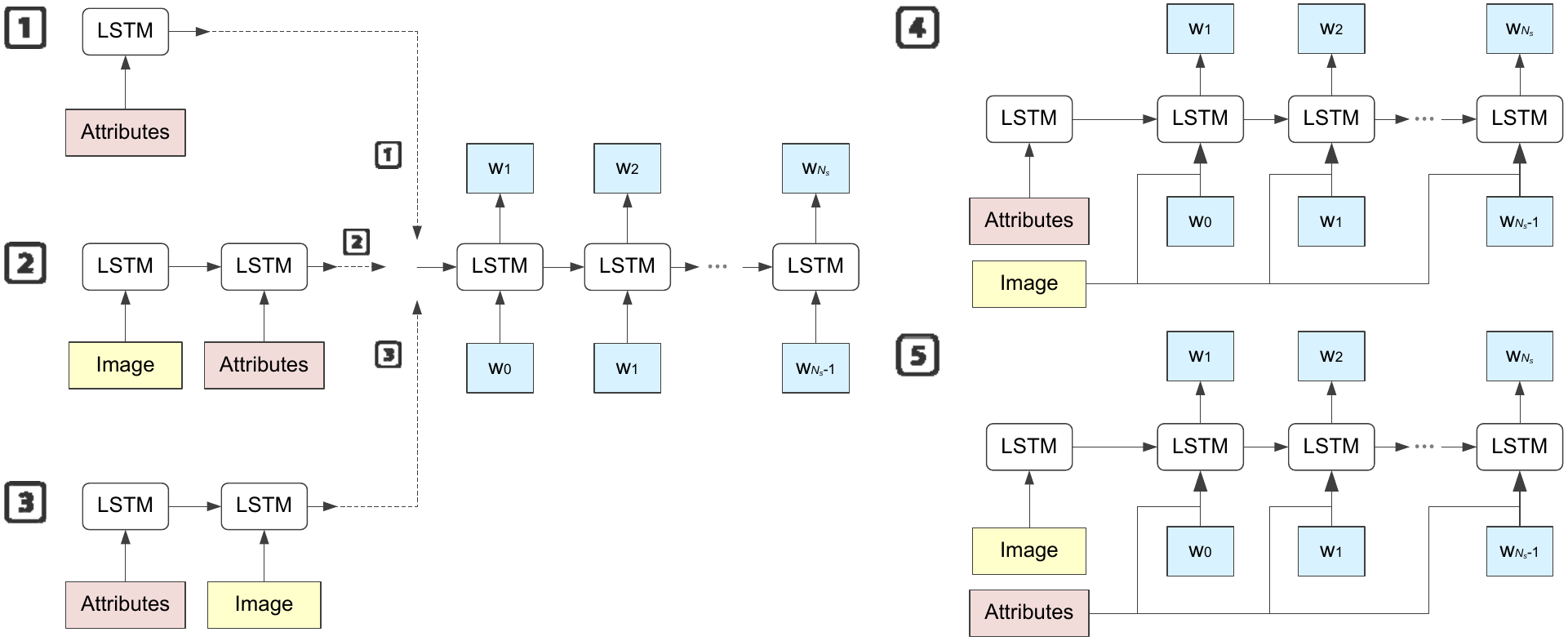}}
\caption{Five variants of our LSTM-A framework (better viewed in color).}
\label{fig:fig1}
\end{figure}

\subsection{Long Short-Term Memory with Attributes}\label{ssec:LSTMA}
Unlike the existing image captioning models in \citep{Donahue14, Vinyals14} which solely encode image representations for sentence generation, our proposed Long Short-Term Memory with Attributes (LSTM-A) model additionally integrates the detected high-level attributes into LSTM. We devise five variants of LSTM-A for involvement of two design purposes. The first purpose is about where to feed attributes into LSTM and three architectures, i.e., {LSTM-A}$_{1}$ (leveraging only attributes), {LSTM-A}$_{2}$ (inserting image representations first) and {LSTM-A}$_{3}$ (feeding attributes first), are derived from this view. The second is about when to input attributes or image representations into LSTM and we design {LSTM-A}$ _{4}$ (inputting image representations at each time step) and {LSTM-A}$_{5}$ (inputting attributes at each time step) for this purpose. An overview of LSTM-A architectures is depicted in Figure \ref{fig:fig1}.

\subsubsection{{LSTM-A}$_{1}$ (Leveraging only Attributes)}
Given the detected attributes, one natural way is to directly inject the attributes as representations at the initial time to inform the LSTM about the high-level attributes. This kind of architecture with only attributes input is named as {LSTM-A}$_{1}$. It is also worth noting that the attributes-based model in \citep{Wu:CVPR16} is similar to {LSTM-A}$_{1}$ and can be regarded as one special case of our LSTM-A. Given the attribute representations ${\bf{A}}$ and the corresponding sentence ${\bf{W}}\equiv [{\bf{w}}_0, {\bf{w}}_1,...,{\bf{w}}_{N_s}]$, the LSTM updating procedure in {LSTM-A}$_{1}$ is as
\begin{eqnarray*}\small
{{\bf{x}}^{-1}} = {{\bf{T}}_{a}}{\bf{A}},~~~~~~~~~~~~~~~~~~~~~~~~~~~~~~~~~~~~~~~~~~~~~~~~~~~~~~~~~~~~\\
{{\bf{x}}^t} = {{\bf{T}}_s}{{\bf{w}}_t},~~t \in \left\{ {0, \ldots ,{N_s}-1} \right\}~~~{\rm{and}}~~~~{{\bf{h}}^{t}} = f\left( {{{\bf{x}}^t}} \right),~~t \in \left\{ {0, \ldots ,{N_s}-1} \right\},
\end{eqnarray*}
where $D_e$ is the dimensionality of LSTM input, ${{\bf{T}}_a} \in {{\mathbb{R}}^{{D_e} \times {D_a}}}$ and ${{\bf{T}}_s} \in {{\mathbb{R}}^{{D_e} \times {D_s}}}$ is the transformation matrix for attribute representations and textual features of word, respectively, and $f$ is the updating function within LSTM unit. Please note that for the input sentence ${\bf{W}} \equiv [ {{{\bf{w}}_0}, \ldots,{{\bf{w}}_{{N_s}}}} ]$, we take ${{\bf{w}}_0}$ as the start sign word to inform the beginning of sentence and ${{\bf{w}}_{{N_s}}}$ as the end sign word which indicates the end of sentence. Both of the special sign words are included in our vocabulary. Most specifically, at the initial time step, the attribute representations are transformed as the input to LSTM, and then in the next steps, word embedding ${{\bf{x}}^t}$ will be input into the LSTM along with the previous step's hidden state ${{\bf{h}}^{t-1}}$. In each time step (except the initial step), we use the LSTM cell output ${{\bf{h}}^{t}}$ to predict the next word. Here a softmax layer is applied after the LSTM layer to produce a probability distribution over all the ${D_s}$ words in the vocabulary as
\begin{equation}\label{Eq:Eq11}\small
{{\Pr}_{t+1}}\left( {{w_{t+1}}} \right) = \frac{{\exp \left\{ {{{\bf{T}}_h^{\left( w_{t+1} \right)}}{{\bf{h}}^t}} \right\}}}{{\sum\limits_{w \in {\mathcal {W}}} {\exp \left\{ {{{{\bf{T}}_h^{\left( w \right)}}}{{\bf{h}}^t} } \right\}} }},
\end{equation}
where $\mathcal {W}$ is the word vocabulary space and ${{\bf{T}}_h^{(w)}}$ is the parameter matrix in softmax layer.

\subsubsection{{LSTM-A}$_{2}$ (inserting image representations first)}
To further leverage both image representations and high-level attributes in the encoding stage of our LSTM-A, we design the second architecture {LSTM-A}$_{2}$ by treating both of them as atoms in the input sequence to LSTM. Specifically, at the initial step, the image representations ${\bf{I}}$ are firstly transformed into LSTM to inform the LSTM about the image content, followed by the attribute representations ${\bf{A}}$ which are encoded into LSTM at the next time step to inform the high-level attributes. Then, LSTM decodes each output word based on previous word ${{\bf{x}}^t}$ and previous step's hidden state ${{\bf{h}}^{t-1}}$, which is similar to the decoding stage in {LSTM-A}$_{1}$. The LSTM updating procedure in {LSTM-A}$_{2}$ is designed as
\begin{eqnarray*}\small
{{\bf{x}}^{-2}} = {{\bf{T}}_{v}}{\bf{I}}~~~{\rm{and}}~~~{{\bf{x}}^{-1}} = {{\bf{T}}_{a}}{\bf{A}},~~~~~~~~~~~~~~~~~~~~~~~~~~~~~~~~~~~~~~~~~~\\
{{\bf{x}}^t} = {{\bf{T}}_s}{{\bf{w}}_t},~~t \in \left\{ {0, \ldots ,{N_s}-1} \right\}~~~{\rm{and}}~~~~{{\bf{h}}^{t}} = f\left( {{{\bf{x}}^t}} \right),~~t \in \left\{ {0, \ldots ,{N_s}-1} \right\},
\end{eqnarray*}
where ${{\bf{T}}_v} \in {{\mathbb{R}}^{{D_e} \times {D_v}}}$ is the transformation matrix for image representations.

\subsubsection{{LSTM-A}$_{3}$ (feeding attributes first)}
The third design {LSTM-A}$_{3}$ is similar to {LSTM-A}$_{2}$ as both designs utilize image representations and high-level attributes to form the input sequence to LSTM in the encoding stage, except that the orders of encoding are different. In {LSTM-A}$_{3}$, the attribute representations are firstly encoded into LSTM and then the image representations are transformed into LSTM at the second time step. The whole LSTM updating procedure in {LSTM-A}$_{3}$ is as
\begin{eqnarray*}\small
{{\bf{x}}^{-2}} = {{\bf{T}}_{a}}{\bf{A}}~~~{\rm{and}}~~~{{\bf{x}}^{-1}} = {{\bf{T}}_{v}}{\bf{I}},~~~~~~~~~~~~~~~~~~~~~~~~~~~~~~~~~~~~~~~~~~~~\\
{{\bf{x}}^t} = {{\bf{T}}_s}{{\bf{w}}_t},~~t \in \left\{ {0, \ldots ,{N_s}-1} \right\}~~~{\rm{and}}~~~~{{\bf{h}}^{t}} = f\left( {{{\bf{x}}^t}} \right),~~t \in \left\{ {0, \ldots ,{N_s}-1} \right\}.
\end{eqnarray*}

\subsubsection{{LSTM-A}$_{4}$ (inputting image representations at each time step)}
Different from the former three designed architectures which mainly inject high-level attributes and image representations at the encoding stage of LSTM, we next modify the decoding stage in our LSTM-A by additionally incorporating image representations or high-level attributes. More specifically, in {LSTM-A}$_{4}$, the attribute representations are injected once at the initial step to inform the LSTM about the high-level attributes, and then image representations are fed at each time step as an extra input to LSTM to emphasize the image content frequently among memory cells in LSTM. Hence, the LSTM updating procedure in {LSTM-A}$_{4}$ is:
\begin{eqnarray*}\small
{{\bf{x}}^{-1}} = {{\bf{T}}_{a}}{\bf{A}},~~~~~~~~~~~~~~~~~~~~~~~~~~~~~~~~~~~~~~~~~~~~~~~~~~~~~~~~~~~~~\\
{{\bf{x}}^t} = {{\bf{T}}_s}{{\bf{w}}_t}+{{\bf{T}}_v}{{\bf{I}}},~~t \in \left\{ {0, \ldots ,{N_s}-1} \right\}~~~{\rm{and}}~~~~{{\bf{h}}^{t}} = f\left( {{{\bf{x}}^t}} \right),~~t \in \left\{ {0, \ldots ,{N_s}-1} \right\}.
\end{eqnarray*}

\subsubsection{{LSTM-A}$_{5}$ (inputting attributes at each time step)}
The last design {LSTM-A}$_{5}$ is similar to {LSTM-A}$_{4}$ except that it firstly encodes image representations and then feeds attribute representations as an additional input to LSTM at each step in decoding stage to emphasize the high-level attributes frequently. Accordingly, the LSTM updating procedure in {LSTM-A}$_{5}$ is as
\begin{eqnarray*}
{{\bf{x}}^{-1}} = {{\bf{T}}_{v}}{\bf{I}},~~~~~~~~~~~~~~~~~~~~~~~~~~~~~~~~~~~~~~~~~~~~~~~~~~~~~~~~~\\
{{\bf{x}}^t} = {{\bf{T}}_s}{{\bf{w}}_t}+{{\bf{T}}_a}{{\bf{A}}},~~t \in \left\{ {0, \ldots ,{N_s}-1} \right\}~~~{\rm{and}}~~~~{{\bf{h}}^{t}} = f\left( {{{\bf{x}}^t}} \right),~~t \in \left\{ {0, \ldots ,{N_s}-1} \right\}.
\end{eqnarray*}

\section{Experiments}\label{sec:EX}
We conducted our experiments on COCO captioning dataset (COCO) \citep{Lin:ECCV14} and evaluated our approaches for image captioning.

\subsection{Dataset}
The dataset, COCO, is the most popular benchmark for image captioning, which contains 82,783 training images and 40,504 validation images. There are 5 human-annotated descriptions per image. As the annotations of the official testing set are not publicly available, we follow the widely used settings in prior works \citep{You:CVPR16,zhou2016image} and take 82,783 images for training, 5,000 for validation and 5,000 for testing.

\subsection{Experimental Settings}
\paragraph{Data Preprocessing.}
Following \citep{Karpathy:CVPR15}, we convert all the descriptions in training set to lower case and discard rare words which occur less than 5 times, resulting in the final vocabulary with 8,791 unique words in COCO dataset.

\paragraph{Features and Parameter Settings.}
Each word in the sentence is represented as ``one-hot" vector (binary index vector in a vocabulary). For image representations, we take the output of 1,024-way $pool5/7\times7\_s1$ layer from GoogleNet \citep{Szegedy14} pre-trained on Imagenet ILSVRC12 dataset \citep{ILSVRC15}. For attribute representations, we select 1,000 most common words on COCO as the high-level attributes and train the attribute detectors with MIL model \citep{Fang:CVPR15} purely on the training data of COCO, resulting in the final 1,000-way vector of probabilities of attributes. The dimensionality of the input and hidden layers in LSTM are both set to 1,024.

\paragraph{Implementation Details.}
We mainly implement our image captioning models based on Caffe \citep{Jia:MM14}, which is one of widely adopted deep learning frameworks. Specifically, with an initial learning rate 0.01 and mini-batch size set 1,024, the objective value can decrease to 25\% of the initial loss and reach a reasonable result after 50,000 iterations (about 123 epochs).

\paragraph{Testing Strategies.}
For sentence generation in testing stage, there are two common strategies. One is to choose the word with maximum probability at each time step and set it as LSTM input for next time step until the end sign word is emitted or the maximum length of sentence is reached. The other strategy is beam search which selects the top-$k$ best sentences at each time step and considers them as the candidates to generate new top-$k$ best sentences at the next time step. We adopt the second strategy and the beam size $k$ is empirically set to 3. 

Moreover, to avoid model-level overfitting, we utilize ensembling strategy to fuse the prediction results of 5 identical models as previous works \citep{Vinyals14,You:CVPR16}. Please note that all the 5 identical models are trained with different initializations separately.

\paragraph{Evaluation Metrics.} For the evaluation of our proposed models, we adopt four metrics: BLEU@$N$ \citep{Papineni:ACL02}, METEOR \citep{Banerjee:ACL05}, ROUGE-L \citep{lin2004rouge}, and CIDEr-D \citep{vedantam2015cider}. All the metrics are computed by using the codes\footnote{\url {https://github.com/tylin/coco-caption}} released by COCO Evaluation Server \citep{chen2015microsoft}.

\subsection{Compared Approaches}
To empirically verify the merit of our LSTM-A models, we compared the following state-of-the-art methods.
\begin{itemize}
  \item NIC \& LSTM \citep{Vinyals14}: NIC attempts to directly translate from image pixels to natural language with a single deep neural network. The image representations are only injected into LSTM at the initial time step. We directly extract the results reported in \citep{You:CVPR16} and name this run as NIC. Furthermore, for fair comparison, we also include one run LSTM which is our implementation of NIC.
  \item LRCN \citep{Donahue14}: LRCN inputs both image representations and previous word into LSTM at each time step for sentence generation.
  \item Hard-Attention \& Soft-Attention \citep{Xu:ICML15}: Spatial attention on convolutional features of an image is incorporated into the encoder-decoder framework through two kinds of mechanisms: 1) ``hard" stochastic attention mechanism equivalently by reinforce learning (Hard-Attention) and 2) ``soft" deterministic attention mechanism with standard back-propagation (Soft-Attention).
  \item ATT \citep{You:CVPR16}: ATT utilizes attributes as semantic attention to combine image representations and attributes in RNN for image captioning.
  \item Sentence-Condition \citep{zhou2016image}: Sentence-condition is proposed most recently and exploits text-conditional semantic attention to generate semantic guidance for sentence generation by conditioning image features on current text content.
  \item MSR Captivator \citep{devlin2015language}: MSR Captivator employs both Multimodal Recurrent Neural Network (MRNN) and Maximum Entropy Language Model (MELM) \citep{Fang:CVPR15} for sentence generation. Deep Multimodal Similarity Model (DMSM) \citep{Fang:CVPR15} is further exploited for sentence re-ranking.
  \item CaptionBot \citep{Tran2016rich}: CaptionBot is a publicly image captioning system\footnote{\label{fn:CB}\url {https://www.captionbot.ai}} which is mainly built on vision models by using Deep residual networks (ResNets) \citep{He:CVPR16} to detect visual concepts, MELM \citep{Fang:CVPR15} language model for sentence generation and DMSM \citep{Fang:CVPR15} for caption ranking. Entity recognition model for celebrities and landmarks is further incorporated to enrich captions and the confidence scoring model is finally utilized to select the output caption.
  \item LSTM-A: {LSTM-A}$_{1}$, {LSTM-A}$_{2}$, {LSTM-A}$_{3}$, {LSTM-A}$_{4}$, and {LSTM-A}$_{5}$ are five variants derived from our proposed LSTM-A framework.
\end{itemize}

\subsection{Performance Comparison}
\paragraph{Performance on COCO}
Table \ref{table:coco} shows the performances of different models on COCO image captioning dataset. Overall, the results across seven evaluation metrics consistently indicate that our proposed LSTM-A exhibits better performance than all the state-of-the-art techniques including non-attention models (NIC, LSTM, LRCN) and attention-based methods (Hard-Attention, Soft-Attention, ATT, Sentence-Condition). In particular, the CIDEr-D can achieve 98.6\%, which is to date the highest performance reported on COCO dataset when extracting image representations by GoogleNet. {LSTM-A}$_{1}$ inputting only high-level attributes as representations makes the relative improvement over LSTM which feeds into image representations instead by 11.6\%, 7.8\%, 5.1\%, and 13.9\% in BLEU@$4$, METEOR, ROUGR-L, and CIDEr-D, respectively. The results basically indicate the advantage of exploiting high-level attributes than image representations for image captioning. Furthermore, by additionally incorporating attributes to LSTM model, {LSTM-A}$_{2}$, {LSTM-A}$_{3}$ and {LSTM-A}$_{5}$ lead to a performance boost, indicating that image representations and attributes are complementary and thus have mutual reinforcement for image captioning. Similar in spirit, {LSTM-A}$_{4}$ improves LRCN by further taking attributes into account. There is a significant performance gap between ATT and {LSTM-A}$_{5}$. Though both runs involve the utilization of image representations and attributes, they are fundamentally different in the way that the performance of ATT is as a result of modulating the strength of attention on attributes to the previous words, and {LSTM-A}$_{5}$ is by employing attributes as auxiliary knowledge to complement image representations. This somewhat reveals the weakness of semantic attention model, where the prediction errors will accumulate quickly along the generated sequence.

\begin{table*}\small
\centering
\caption{\small Performance of our proposed models and other state-of-the-art methods on COCO, where B@$N$, M, R, and C are short for BLEU@$N$, METEOR, ROUGE-L, and CIDEr-D scores. All values are reported as percentage (\%).}
\label{table:coco}
\begin{tabular}{l|c|c|c|c|c|c|c}\hline
~~\textbf{Model}&~~\textbf{B@1}~~&~~\textbf{B@2}~~&~~\textbf{B@3}~~&~~\textbf{B@4}~~
&~~\textbf{M}~~&~~\textbf{R}~~&~~\textbf{C}~~\\ \hline
~~\textbf{NIC} \citep{Vinyals14}     &66.6 & 45.1 & 30.4 & 20.3 & - & - & - \\
~~\textbf{LRCN} \citep{Donahue14}           &62.8 & 44.2 & 30.4 & 21 & - & - & -  \\
~~\textbf{Hard-Attention} \citep{Xu:ICML15} &71.8 & 50.4 & 35.7 & 25 & 23 &-  &- \\
~~\textbf{Soft-Attention} \citep{Xu:ICML15} &70.7 & 49.2 & 34.4 & 24.3 & 23.9 &-  &- \\
~~\textbf{ATT} \citep{You:CVPR16}       &70.9 & 53.7 & 40.2 & 30.4 & 24.3 &-  &- \\
~~\textbf{Sentence-Condition} \citep{zhou2016image} &72 & 54.6 & 40.4 & 29.8 & 24.5 &-  &95.9 \\\hline
~~\textbf{LSTM} \citep{Vinyals14} &68.4  &51.2                &38       &28.4    &23.1  &50.7  &84.3           \\
~~\textbf{{LSTM-A}$_{1}$}  &72.3  &55.8             &42       &31.7    &24.9  &53.3  &96               \\
~~\textbf{{LSTM-A}$_{2}$} &72.8  &56.4              &42.7   &32.2    &25  &53.5  &97.5               \\
~~\textbf{{LSTM-A}$_{3}$} &\textbf{73.1}  &56.4            &42.6   &32.1    &\textbf{25.2}  &53.7  &98.4   \\
~~\textbf{{LSTM-A}$_{4}$}&71.1  &54.5               &40.9   &30.6    &24  &52.5  &90.6   \\
~~\textbf{{LSTM-A}$_{5}$}&73  &\textbf{56.5} &\textbf{42.9} &\textbf{32.5} &25.1  &\textbf{53.8}  &\textbf{98.6}   \\\hline
\end{tabular}
\end{table*}

\begin{table*}\tiny
\centering
\caption{\small Leaderboard of the published state-of-the-art image captioning models on the online COCO testing server (\url{http://mscoco.org/dataset/\#captions-leaderboard}), where B@$N$, M, R, and C are short for BLEU@$N$, METEOR, ROUGE-L, and CIDEr-D scores. All values are reported as percentage (\%).}
\label{table:leaderboard}
\begin{tabular}{l|*{13}{@{~~~}c@{~~~}|}@{~~~}c@{~~~}}\hline
\multicolumn{1}{c|@{~~~}}{\multirow{2}{*}{\textbf{Model}}} & \multicolumn{2}{c|@{~~~}}{\textbf{B@1}} & \multicolumn{2}{c|@{~~~}}{\textbf{B@2}} & \multicolumn{2}{c|@{~~~}}{\textbf{B@3}} & \multicolumn{2}{c|@{~~~}}{\textbf{B@4}} & \multicolumn{2}{c|@{~~~}}{\textbf{M}} & \multicolumn{2}{c|@{~~~}}{\textbf{R}} & \multicolumn{2}{c@{~~~}}{\textbf{C}} \\\cline{2-15}
\multicolumn{1}{c|@{~~~}}{}&c5 &c40 &c5 &c40 &c5 &c40&c5 &c40&c5 &c40&c5 &c40&c5 &c40 \\\hline
\textbf{MSM@MSRA ({LSTM-A}$_{3}$ )}&  \textbf{73.9}  &  \textbf{91.9} &  \textbf{57.5} & \textbf{84.2} &  \textbf{43.6}  &  \textbf{74} & \textbf{33} & \textbf{63.2} & \textbf{25.6} & \textbf{35} & \textbf{54.2} & \textbf{70}  & \textbf{98.4} & \textbf{100.3} \\\hline
\textbf{ATT} \citep{You:CVPR16}&  73.1  &  90 &  56.5 & 81.5 &  42.4  & 70.9 &  31.6 & 59.9 & 25&  33.5 & 53.5& 68.2  & 94.3 &95.8 \\\hline
\textbf{Google} \citep{Vinyals14}&  71.3  &  89.5 &  54.2 & 80.2 &  40.7  & 69.4 &  30.9 & 58.7 & 25.4 &  34.6 & 53& 68.2  & 94.3 &94.6 \\\hline
\textbf{MSR Captivator} \citep{devlin2015language}&  71.5  &  90.7 &  54.3 & 81.9 &  40.7  & 71 &  30.8 & 60.1 & 24.8 &  33.9 & 52.6 & 68  & 93.1 & 93.7 \\\hline
\end{tabular}
\end{table*}

Compared to {LSTM-A}$_{1}$, {LSTM-A}$_{2}$ which is augmented by integrating image representations performs better, but the performances are lower than {LSTM-A}$_{3}$. The results indicate that {LSTM-A}$_{3}$, in comparison, is benefited from the mechanism of first feeding attributes into LSTM instead of starting from inserting image representations in {LSTM-A}$_{2}$. The chance that a good start point can be attained and lead to performance gain is better. {LSTM-A}$_{4}$ feeding the image representations at each time step yields inferior performances to {LSTM-A}$_{3}$, which only inputs image representations once. We speculate that this may because the noise in the image can be explicitly accumulated and thus the network overfits more easily. In contrast, the performances of {LSTM-A}$_{5}$ which feeds attributes at each time step show the improvements on {LSTM-A}$_{3}$. The results demonstrate that the high-level attributes are more accurate and easily translated into human understandable sentence. Among the five proposed LSTM-A architectures, {LSTM-A}$_{3}$ achieves the best performances in terms of BLEU@1 and METEOR, while {LSTM-A}$_{5}$ performs the best in other five evaluation metrics.

\begin{figure*}
\centering {\includegraphics[width=0.95\textwidth]{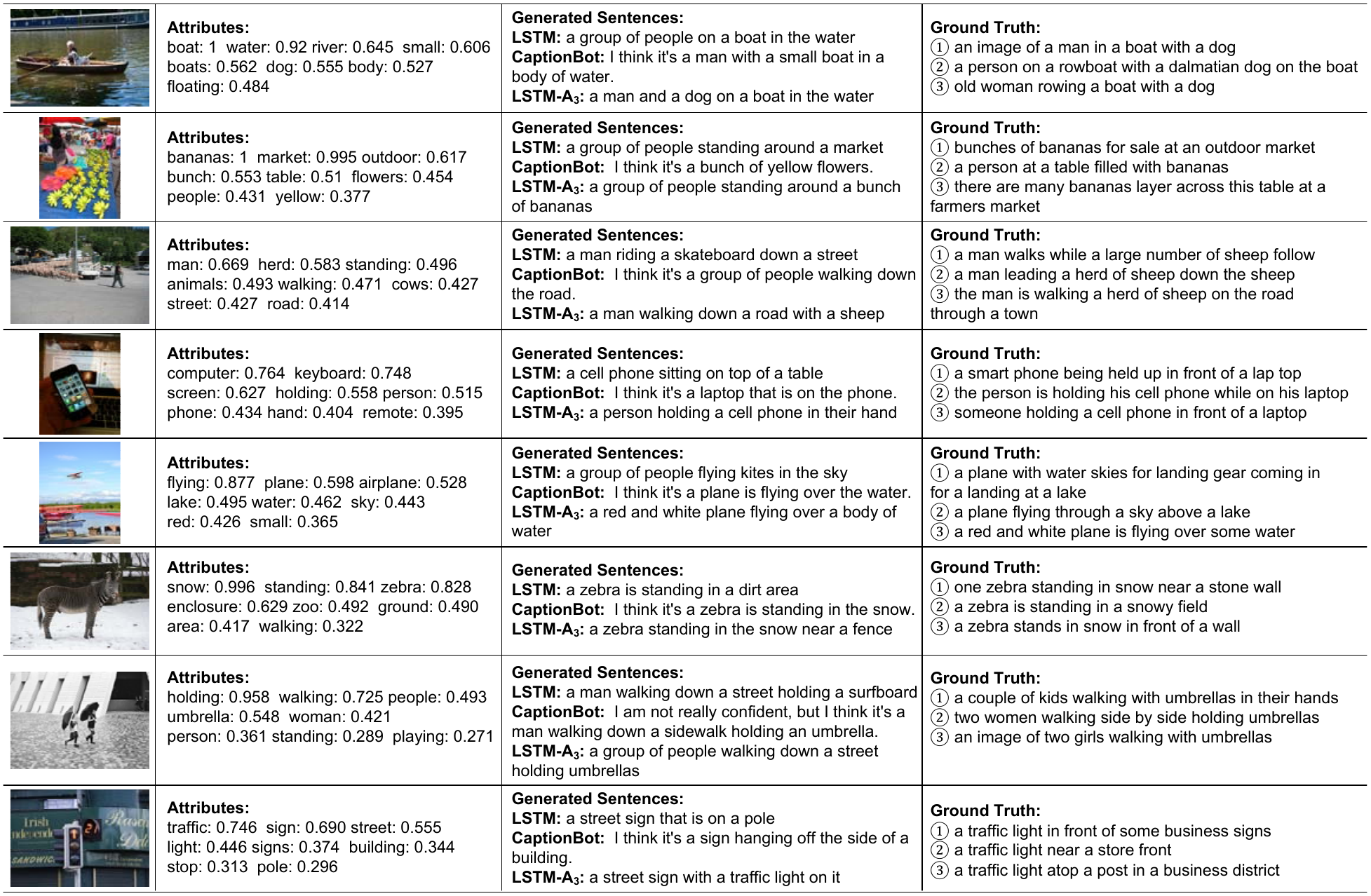}}
\caption{\small Attributes and sentences generation results on COCO. The attributes are predicted by MIL method in \citep{Fang:CVPR15} and the output sentences are generated by 1) LSTM, 2) CaptionBot\textsuperscript{\ref{fn:CB}}, 3) our {LSTM-A}$_{3}$, and 4) Ground Truth: randomly selected three ground truth sentences.}
\label{fig:figs_example}
\vspace{-0.12in}
\end{figure*}

\paragraph{Performance on COCO online testing server}
We also submitted our best run in terms of METEOR, i.e., {LSTM-A}$_{3}$, to online COCO testing server and evaluated the performance on official testing set. Table \ref{table:leaderboard} shows the performance Leaderboard on official testing image set with 5 reference captions (c5) and 40 reference captions (c40). Please note that here we utilize the outputs of 2,048-way $pool5$ layer from ResNet-152 as image representations in our final submission and only the latest top-3 performing methods which have been officially published are included in the table. Compared to the top performing methods, our proposed {LSTM-A}$_{3}$ achieves the best performance across all the evaluation metrics on both c5 and c40 testing sets, and to-date ranks the first on the Leaderboard. In addition, when training the attribute detectors by ResNet-152, our CIDEr-D scores on c5 and c40 testing sets will be further boosted up to 104.9\% and 105.3\%, respectively.

\begin{figure*}
\centering
\caption{The effect of beam size $k$ on (a) {LSTM-A}$_{3}$ and (b) {LSTM-A}$_{5}$.}
\label{fig:beamsize}
\subfigure[$k$ for {LSTM-A}$_{3}$]{
{\includegraphics[width=0.4\textwidth]{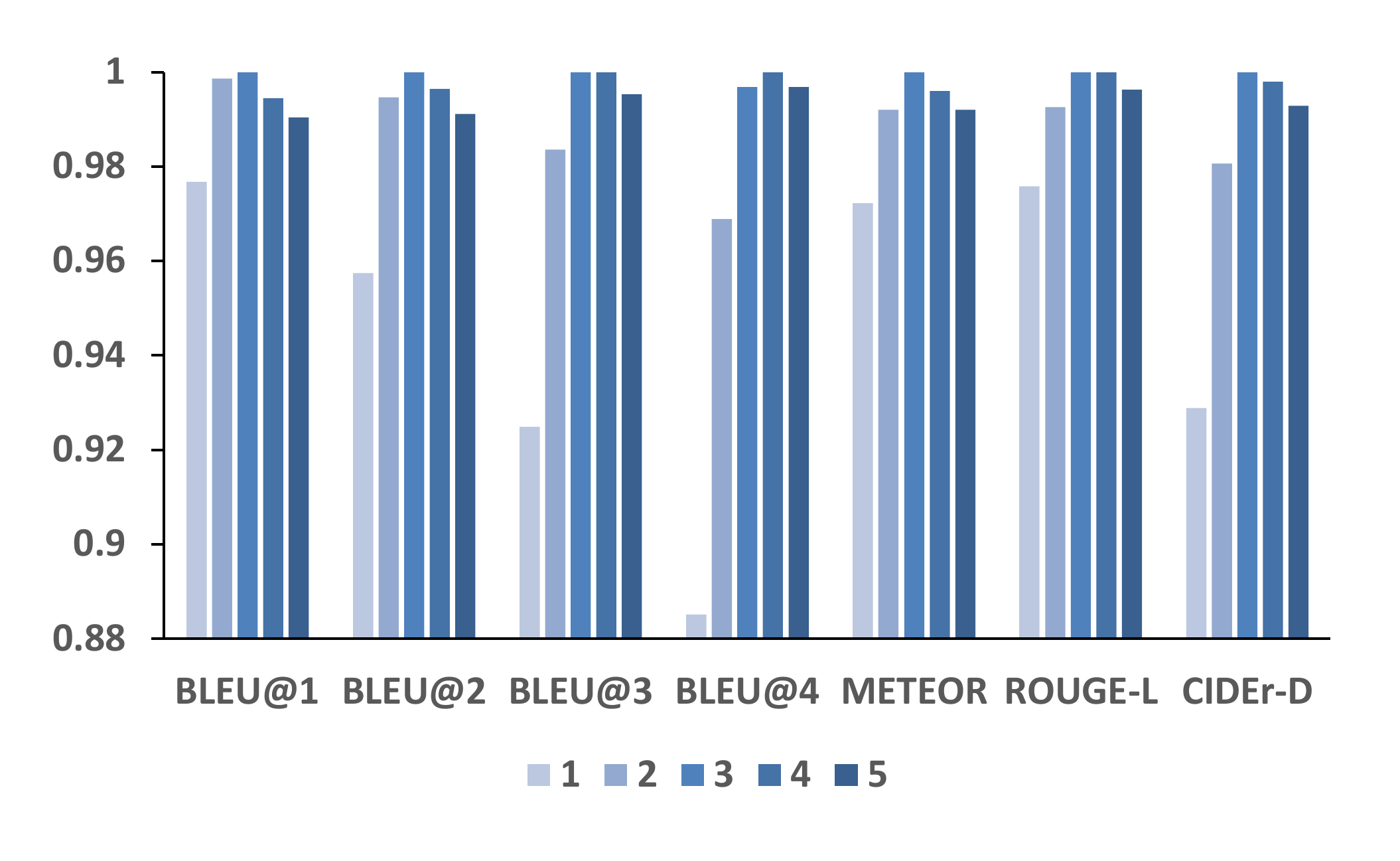}}
}
\subfigure[$k$ for {LSTM-A}$_{5}$]{
{\includegraphics[width=0.4\textwidth]{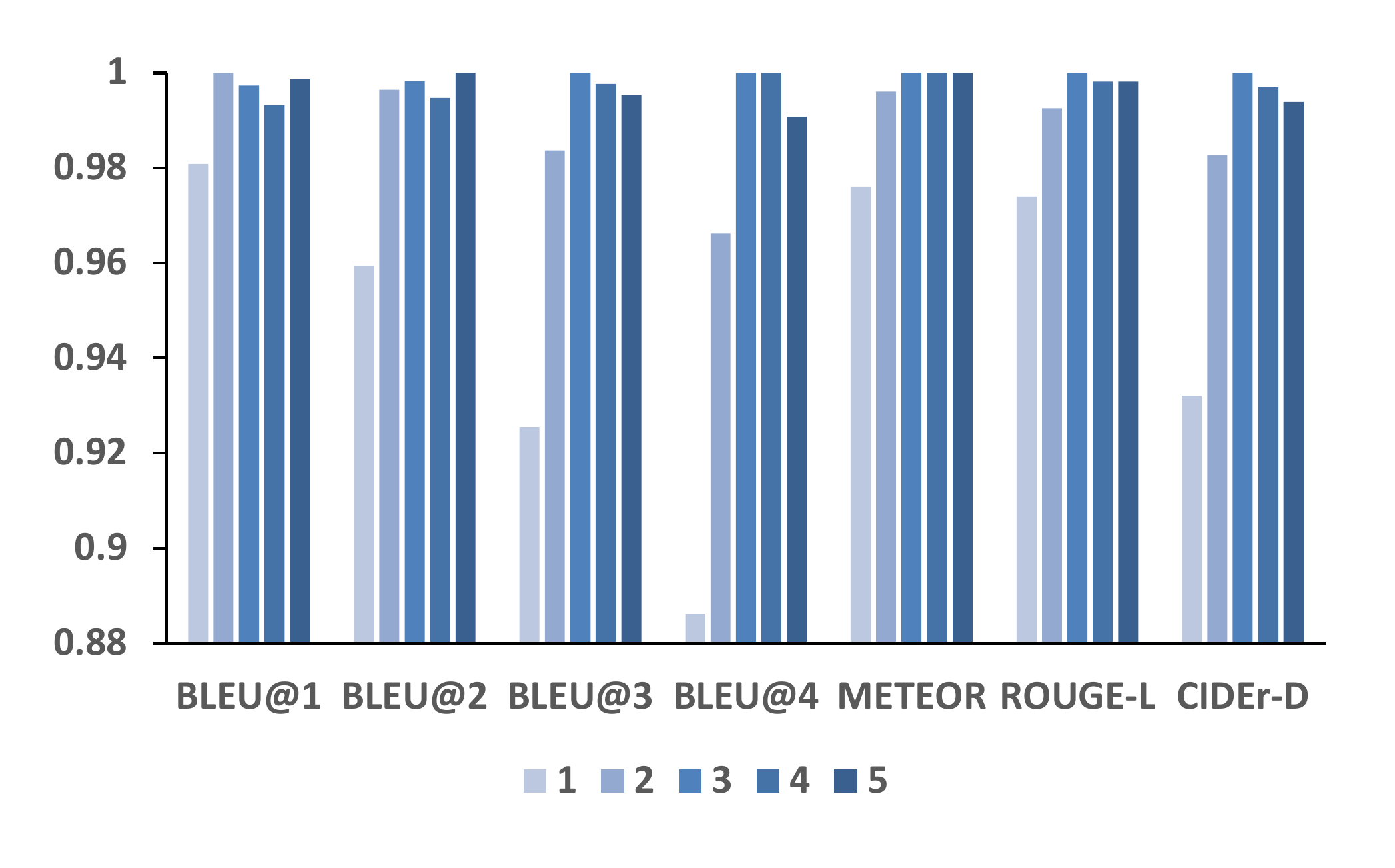}}
}
\end{figure*}

\subsection{Qualitative Analysis}
Figure \ref{fig:figs_example} shows a few sentence examples generated by different methods, the detected high-level attributes, and human-annotated ground truth sentences. From these exemplar results, it is easy to see that all of these automatic methods can generate somewhat relevant sentences, while our proposed {LSTM-A}$_{3}$ can predict more relevant keywords by jointly exploiting high-level attributes and image representations for image captioning. For example, compared to subject term ``a group of people" and ``a man" in the sentence generated by LSTM and CaptionBot respectively, ``a man and a dog" in our {LSTM-A}$_{3}$ is more precise to describe the image content in the first image, since the keyword ``dog" is one of the detected attributes and directly injected into LSTM to guide the sentence generation. Similarly, verb term ``holding" which is also detected as one high-level attribute presents the fourth image more exactly. Moreover, our {LSTM-A}$_{3}$ can generate more descriptive sentence by enriching the semantics with high-level attributes. For instance, with the detected adjective ``red," the generated sentence ``a red and white plane flying over a body of water" of the fifth image depicts the image content more comprehensive.

\subsection{Analysis of the Beam Size $k$}\label{sssec:BS}
In order to analyze the effect of the beam size $k$ in testing stage, we illustrate the performances of our two top performing architectures {LSTM-A}$_{3}$ and {LSTM-A}$_{5}$ with the beam size in the range of \{1, 2, 3, 4, 5\} in Figure \ref{fig:beamsize}. To make all performances fall into a comparable scale, all scores are normalized by the highest score of each evaluation metric. As shown in Figure \ref{fig:beamsize}, we can see that almost all performances in terms of each evaluation metric are like the ``$\wedge$" shapes when beam size $k$ varies from 1 to 5. Hence, we set the beam size $k$ as 3 in our experiments, which can achieve the best performance with a relatively small beam size.

\section{Discussions and Conclusions}\label{sec:CON}
We have presented Long Short-Term Memory with Attributes (LSTM-A) architectures which explores both image representations and high-level attributes for image captioning. Particularly, we study the problem of augmenting high-level attributes from images to complement image representations for enhancing sentence generation. To verify our claim, we have devised variants of architectures by modifying the placement and moment, where and when to feed into the two kinds of representations. Experiments conducted on COCO image captioning dataset validate our proposal and analysis. Performance improvements are clearly observed when comparing to other captioning techniques and more remarkably, the performance of our LSTM-A to date ranks the first on COCO image captioning Leaderboard.

Our future works are as follows. First, more attributes will be learnt from large-scale image benchmarks, e.g., YFCC-100M dataset, and integrated into image captioning. We will further analyze the impact of different number of attributes involved. Second, how to enlarge the word vocabulary of generated sentences with the learnt attributes is worth trying and seems very interesting.

\small
\bibliography{iclr2017_conference}
\bibliographystyle{iclr2017_conference}

\end{document}